\documentclass[twocolumn]{cinc}

\usepackage{amsmath, graphicx, xurl, hyperref, amssymb, comment, booktabs, tabularx, multirow, rotating, adjustbox, bbold}
\usepackage[dvipsnames]{xcolor}
\newcolumntype{L}{>{\raggedright\arraybackslash}X}
\newcolumntype{C}{>{\centering\arraybackslash}X}

\begin{document}

\title{Lirot.ai: A Novel Platform for Crowd-Sourcing Retinal Image Segmentations}


\author {Jonathan Fhima$^{1,2}$, Jan Van Eijgen$^{3,4}$, Moti Freiman$^{1}$, Ingeborg Stalmans$^{3,4}$,  Joachim A. Behar$^{1}$  \\
\ \\ 
 $^1$Faculty of Biomedical Engineering, Technion-IIT, Haifa, Israel \\
 $^2$Department of Applied Mathematics Technion–IIT, Haifa, Israel\\
 $^3$ Research Group Ophthalmology, Department of Neurosciences, KU Leuven, Belgium \\
 $^4$ Department of Ophthalmology, University Hospitals UZ Leuven, Belgium}

\maketitle
\begin{abstract}

\textbf{Introduction}: For supervised deep learning (DL) tasks, researchers need a large annotated dataset. In medical data science, one of the major limitations to develop DL models is the lack of annotated examples in large quantity. This is most often due to the time and expertise required to annotate. We introduce Lirot.ai, a novel platform for facilitating and crowd-sourcing image segmentations.  \textbf{Methods}: Lirot.ai is composed of three components; an iPadOS client application named Lirot.ai-app, a backend server named Lirot.ai-server and a python API name Lirot.ai-API. Lirot.ai-app was developed in Swift 5.6 and Lirot.ai-server is a firebase backend. Lirot.ai-API allows the management of the database. Lirot.ai-app can be installed on as many iPadOS devices as needed so that annotators may be able to perform their segmentation simultaneously and remotely. We incorporate Apple Pencil compatibility, making the segmentation faster, more accurate, and more intuitive for the expert than any other computer-based alternative. \textbf{Results}: We demonstrate the usage of Lirot.ai for the creation of a retinal fundus dataset with reference vasculature segmentations. \textbf{Discussion and future work}: We will use active learning strategies to continue enlarging our retinal fundus dataset by including a more efficient process to select the images to be annotated and distribute them to annotators.
\end{abstract}

\section{Introduction}

Data annotation is the starting point for supervised learning research. Thus the elaboration of a large annotated dataset is critical \cite{Cho2015HowAccuracy} for training a supervised DL model and reaching performances that may enable translational research to the clinical practice. Nevertheless, data annotation is time-consuming, and researchers often require expert annotators. In that respect, the lack of quality reference annotations has often become a major limitation for research \cite{Mamoshina2016ApplicationsBiomedicine}. Publication including the acquisition of reference segmentation do not systematically describe the software used to support the annotation process. Other works refer to computer-based software making use of the keyboard and mouse for segmentation (\cite{Sharma2017CrowdsourcingClassification,Sivaswamy2014Drishti-gs:Segmentation}). These software are often non  intuitive for non-initiated annotators, making the segmentation task more tedious.
To counter this limitation, we decided to build a tailor-made platform that optimizes the segmentation process. The new platform, called Lirot.ai, is composed of three components; an iPadOS application named Lirot.ai-app, a firebase \cite{Moroney2017TheDatabase,Moroney2017CloudFirebase} backend server called Lirot.ai-server and a python API named Lirot.ai-API. We used Lirot.ai to create a retinal fundus dataset with arterioles and venules segmentation from digital fundus images (DFI). Lirot.ai aims to: (1) enable simultaneous and remote usage by an unlimited number of annotators, (2) the user interface (UI) should be tailored to segmentation tasks, and (3) facilitates the distribution of the data to the annotators and centralize the segmentations towards research usage.

\section{Methods}
Lirot.ai was built to facilitate the segmentation workload. The researcher can first decide what images need to be segmented, then use the python Lirot.ai-API to transfer these images to the Lirot.ai-server. The annotators will then be notified through the Lirot.ai-app and will be able to download the images, segment them and upload the segmentations to the Lirot.ai-server. All these operations are easily performed using the Lirot.ai-app. Finally, the researcher can download the segmentations from all annotators and use them for training a DL segmentation model.

\subsection{Lirot.ai-app}
\label{app}
\begin{figure*}
\centering
\includegraphics[page=1,width=0.9\textwidth]{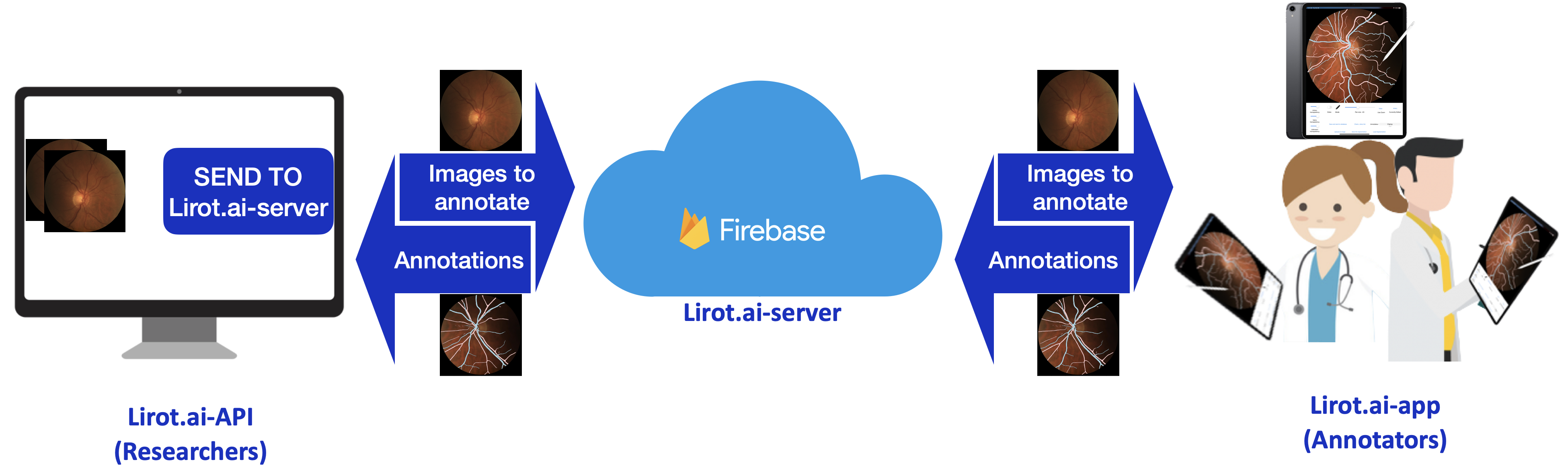}
\caption{Pipeline of the Lirot.ai platform: a dataset to annotate is defined by the research team and uploaded to the Lirot.ai-server. The annotators can download and segment images on their iPad. When finished, they upload the images to the Lirot.ai-server.}
\label{LirotPipe}
\end{figure*}

Lirot.ai-app has been developed in Swift 5.6, the native language for iPadOS development. That enables optimal reactivity and speed using the app. Lirot.ai-app has been elaborated towards facilitating image segmentation and crowd-sourcing. 
Its UI has been designed to optimize the time invested in the manual segmentation process by helping the annotators to perform the segmentation task faster and more accurately.
The application can be installed on iPadOS devices and enables annotators to work simultaneously and remotely. Lirot.ai enables to use an active learning (AL) module that can automatically select what images need to be annotated next. This is made possible by the ease of image transfer from the server to the client application, to distribute the next set of images to be annotated, and from the client app to the sever following the segmentation task, thus facilitating the automated selection of the next batch of images to be annotated based on the updated model performance. Indeed, the annotators can directly download images via Lirot.ai-app, segment them, and send back the segmentation to the Lirot.ai-server without needing any tierce application. The work pipeline in using Lirot.ai is shown in Figure \ref{LirotPipe}.
We also decided to add functionalities to reduce the time required to annotate an image and maximize the segmentation quality. This includes a pre-segmentation option that can provide a rough, automatically generated segmentation that the annotator will have to correct instead of starting from scratch. It also includes an image quality assessment score automatically generated using FundusQ-Net \cite{AbramovichFundusQ-Net:Grading}. Furthermore, the researcher has access to a version control manager to see the history of all the segmentations sent to Lirot.ai-server. Thus, it is possible to restore a previous version of the segmentation for a given image in case of a wrong segmentation was sent to Lirot.ai-server by mistake.

We developed the UI of Lirot.ai-app (Figure \ref{Lirot}) to be more intuitive than the current computer-based alternative. We incorporated the Apple Pencil compatibility, transforming the segmentation task into a simple pencil-based drawing task. More advanced features have also been implemented to improve the trade-off between the quality of the segmentation and the time invested in this task; we included inside Lirot.ai-app a change in the transparency of the different segmented classes - which enables segmenting the image while seeing the background -, a contrast enhancer mode - which allows the annotators to have a better view of the darkest part of the images -, a zoom feature and a pencil size that the annotators can adjust. Practically, those features allow the annotators to segment the fine details of the images more quickly and accurately. The contrast enhancer and the zoom feature (Figure \ref{Lirot2}) enable the annotators to see the important part of the image more easily. The contrast enhancer can be applied by clicking on the related button, and the zoom feature can be used easily by pinching the image with two fingers. The pencil size enables the user to have a better time-quality trade-off by using a small pencil size to annotate fine details and a bigger pencil size otherwise.

\subsection{Active Learning}
AL is the subset of machine learning in which a learning algorithm can query a user interactively to annotate data with the desired outputs. In AL, the algorithm proactively selects the subset of examples to be annotated next from the pool of non-annotated data. AL is a known procedure to improve the model’s generalization capability if the selected strategy is well defined for the task. 
Xie et al. \cite{Xie2020Deal:Segmentation} succeeded to achieve the maximal performance on the segmentation of the cityscapes dataset \cite{Cordts2016TheUnderstanding} by using only $40\%$ of it, selected by AL. The working pipeline of Lirot.ai is designed to allow the researchers to easily implement their active learning strategies by facilitating the transfer of data between the annotators and the researchers.

\subsection{Pre-segmentation}
In order to help the annotators to perform a faster manual segmentation, Lirot.ai can be used with any DL-based TensorFlow Lite model embedded for pre-segmentation. This enables the annotators to accelerate the segmentation task by starting from the pre-segmented image generated by a baseline DL model (Figure \ref{Lirot2}). 

\subsection{Image quality assessment}
The quality of an image can sometimes be too low to be exploitable and should be discarded. We have included the possibility to incorporate a TensorflowLite DL-based model trained to evaluate the quality of the images inside Lirot.ai-app. It will provide the annotators with a quality grade for a given image. If the quality of an image is too low, the annotator can decide to skip the image. 

\subsection{Quality control}
Another critical aspect of image segmentation is the variability of expertise across annotators. For that reason, the Lirot.ai framework enables to set senior annotators with more expertise who can oversee the segmentation performed by all annotators and eventually adjust/correct them if necessary. These senior annotators have a special version of the Lirot.ai-app where they have access to the history of all the segmentations sent by their coworkers. They are able to correct segmentations as needed and upload the corrected segmentations to Lirot.ai-server. Lirot.ai-server will log if a senior annotator has double-checked and/or modified a segmentation. Lirot.ai also includes a version control manager that enables the researchers to access all the previous versions of the segmenations uploaded toLirot.ai-server, including the timestamp and annotator name. This can be useful, for example, in the scenario where a previous version of the segmentation needs to be retried.

\subsection{Lirot.ai-server}
Lirot.ai-server uses both firebase realtime database \cite{Moroney2017TheDatabase} and firebase cloud storage \cite{Moroney2017CloudFirebase} in order to manage the backend of Lirot.ai. Firebase is a solution provided by Google to facilitate the development of applications. Firebase realtime database allows us to save text information about our images, segmentations, and annotators. We use it to save the information about the annotators, the timestamp of the segmentation, the history of the previous segmentation sent for each image, and the link of the document we stored inside firebase cloud storage. Firebase cloud storage is a cloud solution that allows firebase users to save documents. We use it to save the images that the researchers want to send to the annotators and the segmentations that the annotators want to send back to the researchers.

\subsection{Lirot.ai-API}
Lirot.ai-API is a python API which allows the researchers to communicate with Lirot.ai-server. It is used to send images which require segmentation and to download the segmentations made by the annotators.

\subsection{Vasculature segmentation DFI}
\begin{figure}[h]
\centering
\includegraphics[page=0.5,width=0.45\textwidth]{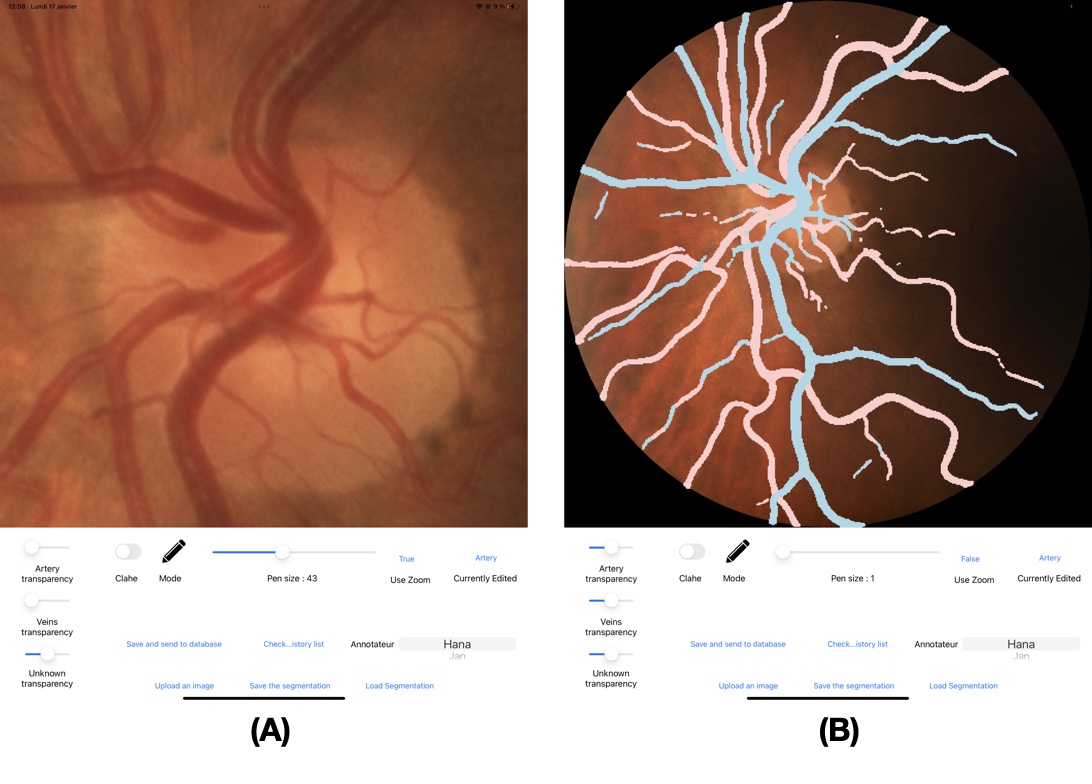}
\caption{User interface of Lirot.ai-app during the segmentation of the arterioles and venules of DFI. (A): Zoomed DFI, (B): Pre-segmentation generated by our baseline model}
\label{Lirot2}
\end{figure}

For our use case, we used Lirot.ai for vasculature segmentation in DFIs. The resolution of the DFIs we used is 1444x1444 pixels. We used Lirot.ai-app on iPad Pro 13" and 11". For the particular task of retinal image quality assessment, we used FundusQ-Net developed by Abramovich et al. \cite{AbramovichFundusQ-Net:Grading} which accurately grades the retinal fundus images quality. We have also developed a custom DL baseline model to achieve pre-segmentation (Figure \ref{Lirot2}). 

\section{Results}
\begin{figure}[h]
\centering
\includegraphics[page=0.5,width=0.5\textwidth]{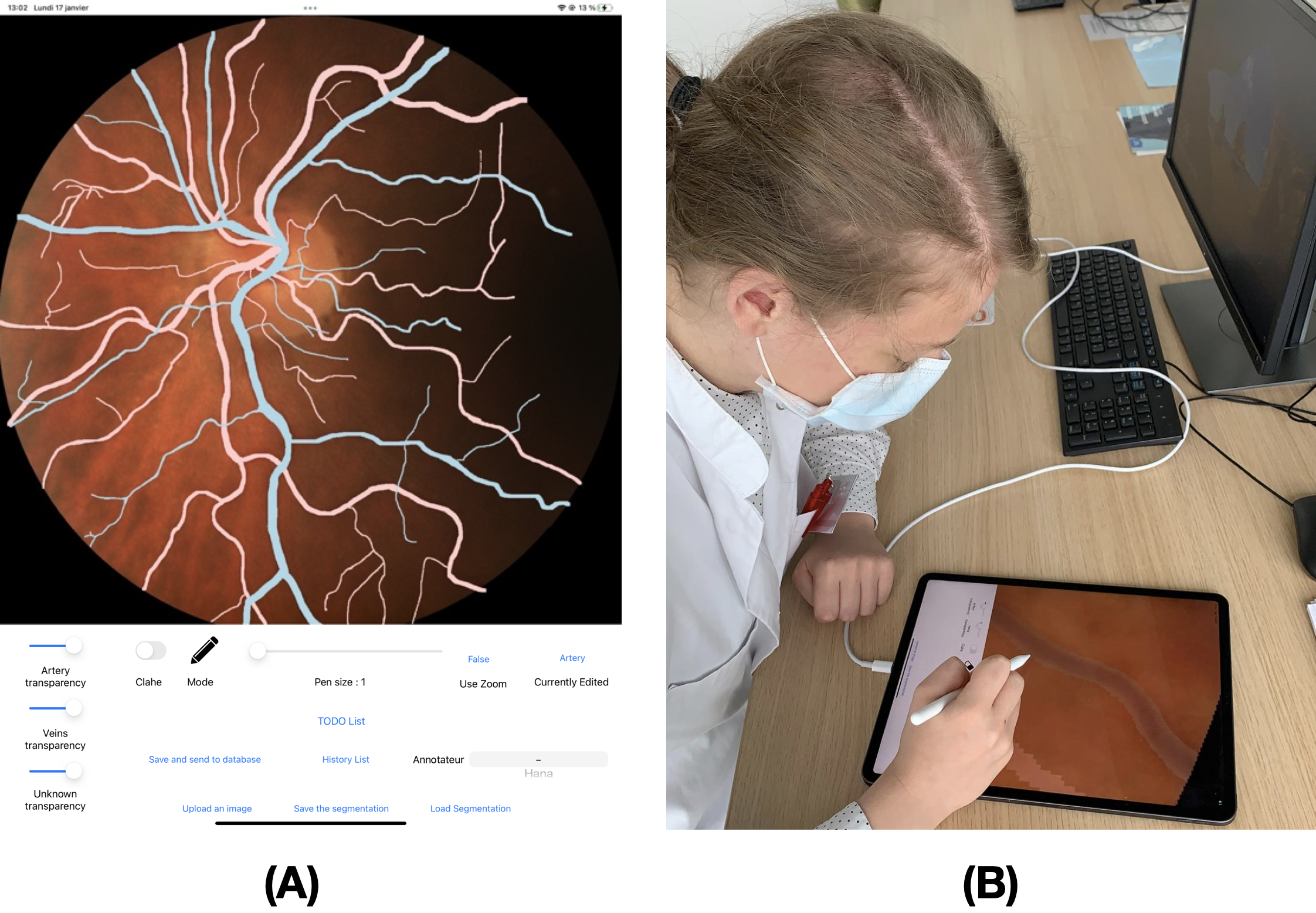}
\caption{(A): User interface of Lirot.ai-app during the segmentation of the arterioles and venules of DFI. (B): Annotator segmenting a DFI with Lirot.ai-app in the UZ Leuven hospital. }
\label{Lirot}
\end{figure}

\begin{figure}[h]
\centering
\includegraphics[page=0.5,width=0.5\textwidth]{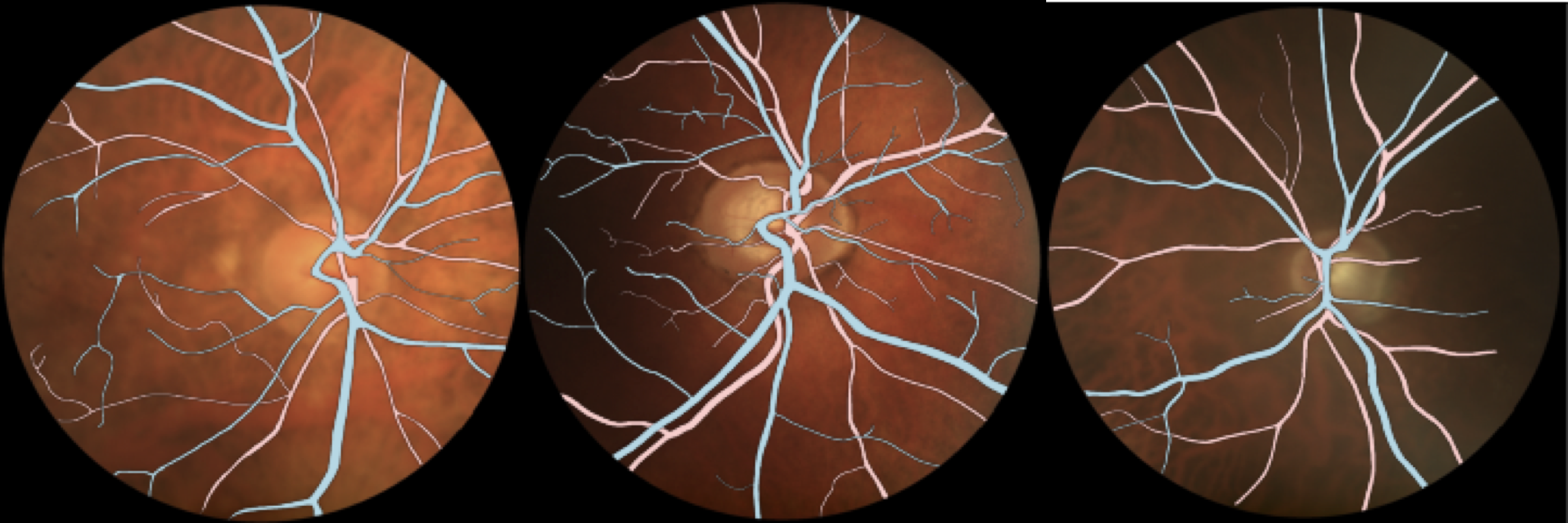}
\caption{Example of generated DFIs segmentation using Lirot.ai}
\label{example}
\end{figure}

Lirot.ai has been used in collaboration with the Universitair Ziekenhuis (UZ) Leuven hospital since 09-2021 to construct a dataset of DFIs with segmented vasculature. A total of four medical students and one expert ophthalmologist (senior annotator) experienced in microvascular research used Lirot.ai-app to segment the DFIs. It enables us to crowd-source the reference segmentations and elaborate our dataset in a more timely manner than using sequential segmentation on a single computer located at the clinic as done previously by others. Another advantage of working with Lirot.ai is the ease of transferring the image to annotate to the annotators and downloading the segmentations. A few examples of the generated dataset can be seen in Figure \ref{example}. Figure \ref{Lirot} shows an annotator segmenting a DFI with Lirot.ai-app in the UZ Leuven hospital.

\section{Discussion and conclusion}
We developed Lirot.ai, a novel platform optimized for crowd-sourcing image segmentations.
It is composed of a python API named Lirot.ai-API, a firebase server named Lirot.ai-server, and an iPadOS application called Lirot.ai-app. The Lirot.ai-app is easy to install and use. It can be used simultaneously by an unlimited number of annotators, which could significantly accelerate the construction of a database.
Lirot.ai-API allows the researchers to manage the database and select the images they want to send to the annotators via the Lirot.ai-server. The annotators can then download and segment the images directly via Lirot.ai-app before sending them back to the Lirot.ai-server.
We have included several tools to reduce the time-quality trade-off required to annotate an image, including an intuitive UI, a pre-segmentation option, an image quality assessment option, and a quality control interface to simplify the work of the annotators without losing segmentation quality.
We have also included all the technology required to develop an AL pipeline to optimize the selection of the next images to segment. We used Lirot.ai to create a DFI dataset with reference vasculature segmentations. In future work, we aim to use an AL framework to continue enlarging our database more efficiently than a random sampling of the next images to annotate.

\section*{Acknowledgments}  
JF, JB, and MF were supported by Grant No ERANET - 2031470 from the Ministry of Health.

\bibliography{template}

\begin{correspondence}
Jonathan Fhima\\
jonathanfh@campus.technion.ac.il
\end{correspondence}

\end{document}